\documentclass[11pt]{article}
\pdfoutput=1
\usepackage[
  margin=0.8in,
  headheight=12pt,
  headsep=25pt,
  includefoot,
  footskip=30pt,
]{geometry}
\counterwithin{figure}{section}
\usepackage{tikz}
\usepackage{listings}
\usepackage{xcolor}
\usepackage{inconsolata}
\definecolor{commentcolour}{rgb}{0.3,0.7,0.2}
\definecolor{backcolour}{rgb}{0.98,0.98,0.98}

\lstdefinelanguage{markdown}{
    comment=[l]{\#},
    morestring=[s]{```}{```},
    commentstyle=\color{commentcolour}\bfseries,
    stringstyle=\color{blue},
    basicstyle=\scriptsize\ttfamily,
    showstringspaces=false,
    breaklines=true,
    breakautoindent=false,
    breakindent=0pt,
    backgroundcolor=\color{backcolour},
}
\lstdefinestyle{mystyle}{
    morekeywords={self},
    basicstyle=\scriptsize\ttfamily,
    keywordstyle=\color{blue},
    commentstyle=\color{commentcolour}\bfseries,
    breaklines=true,
    breakautoindent=false,
    showstringspaces=false,
    backgroundcolor=\color{backcolour},
    stringstyle=\color{red},
    escapechar=@
}
\lstdefinelanguage{PythonPlus}[]{Python}{
  morekeywords=[1]{,as,assert,nonlocal,with,yield,self,True,False,None,} %
  morekeywords=[2]{,__init__,__add__,__mul__,__div__,__sub__,__call__,__getitem__,__setitem__,__eq__,__ne__,__nonzero__,__rmul__,__radd__,__repr__,__str__,__get__,__truediv__,__pow__,__name__,__future__,__all__,}, %
  morekeywords=[3]{,object,type,isinstance,copy,deepcopy,zip,enumerate,reversed,list,set,len,dict,tuple,range,xrange,append,execfile,real,imag,reduce,str,repr,}, %
  morekeywords=[4]{,Exception,NameError,IndexError,SyntaxError,TypeError,ValueError,OverflowError,ZeroDivisionError,}, %
  morekeywords=[5]{,ode,fsolve,sqrt,exp,sin,cos,arctan,arctan2,arccos,pi, array,norm,solve,dot,arange,isscalar,max,sum,flatten,shape,reshape,find,any,all,abs,plot,linspace,legend,quad,polyval,polyfit,hstack,concatenate,vstack,column_stack,empty,zeros,ones,rand,vander,grid,pcolor,eig,eigs,eigvals,svd,qr,tan,det,logspace,roll,min,mean,cumsum,cumprod,diff,vectorize,lstsq,cla,eye,xlabel,ylabel,squeeze,}, %
}

\usepackage{tikz}
\usetikzlibrary{shapes.emoticon}

\usepackage{amsmath} %
\usepackage{array} %
\usetikzlibrary{matrix, arrows}
\usepackage{amsmath,amssymb}
\usepackage{amsthm}
\usepackage{mathtools}
\usepackage{xspace}
\usepackage[noend]{algorithmic}
\usepackage[ruled,vlined]{algorithm2e}
\usepackage{url}
\usepackage{makeidx}
\usepackage{enumerate}
\usepackage{epstopdf}
\usepackage{booktabs}
\usepackage{color}
\usepackage[utf8]{inputenc}
\usepackage{thm-restate}
\usepackage{scalerel,stackengine}
\usepackage[shortlabels]{enumitem}
\usepackage{xr}
\usepackage{fancyvrb}
\usepackage{xcolor}
\usepackage{bold-extra}
\usepackage{arydshln}
\usepackage[small]{caption}
\usepackage{subcaption}
\usepackage[most]{tcolorbox}
\usepackage{fvextra}
\usepackage{float}
\usepackage{alltt}
\usepackage{soul}
\usepackage{MnSymbol,wasysym}
\usepackage{fancyvrb}
\usepackage{multirow}
\usepackage[final]{hyperref}
\usepackage[bottom]{footmisc}
\usepackage{diagbox}
\usepackage{mdframed}
\usepackage{todonotes}
\setuptodonotes{inline}
\usepackage{tikz}
\usetikzlibrary{shapes,calc,positioning}

\global

\tcbset{
  aibox/.style={
    width=\textwidth,
    top=0pt, bottom=0pt, left=5pt, right=5pt,
    colback=white,
    colframe=black,
    colbacktitle=black,
    enhanced,
    center,
    attach boxed title to top left={yshift=-0.1in,xshift=0.15in},
    boxed title style={boxrule=0pt,colframe=white,},
  }
}
\newtcolorbox{AIbox}[2][]{aibox,title=#2,#1}

\definecolor{aigold}{RGB}{244,210, 1} 
\definecolor{aigreen}{RGB}{210,244,211} 

\sethlcolor{aigreen}

\definecolor{aired}{RGB}{255,180,181} 
\newcommand{\lightred}[1]{\colorbox{aired}{\parbox{\linewidth}{#1}}}

\newtcbox{\mybox}[1][green]{on line,
arc=0pt,outer arc=0pt,colback=#1!10!white,colframe=#1!50!black,
boxsep=0pt,left=0pt,right=0pt,top=0pt,bottom=0pt,
boxrule=0pt,bottomrule=0pt,toprule=0pt}

\newcommand{\phione}{\textbf{phi-1} }
\newcommand{\phionesmall}{\textbf{phi-1-{small}} }
\newcommand{\phionebase}{\textbf{phi-1-{base}} }

\begin{document}

\title{Textbooks Are All You Need}

\author{Suriya Gunasekar
\and Yi Zhang
\and Jyoti Aneja
\and Caio C\'esar Teodoro Mendes\
\and Allie Del Giorno
\and Sivakanth Gopi
\and Mojan Javaheripi
\and Piero Kauffmann
\and Gustavo de Rosa
\and Olli Saarikivi
\and Adil Salim
\and Shital Shah
\and Harkirat Singh Behl
\and Xin Wang
\and S\'ebastien Bubeck
\and Ronen Eldan
\and Adam Tauman Kalai
\and Yin Tat Lee
\and Yuanzhi Li}

\date{Microsoft Research}

\maketitle

\begin{abstract}
We introduce \textbf{phi-1}, a new large language model for code, with significantly smaller size than competing models: \textbf{phi-1} is a Transformer-based model with $1.3$B parameters, trained for $4$ days on $8$ A100s, using a selection of ``textbook quality" data from the web ($6$B tokens) and synthetically generated textbooks and exercises with GPT-3.5 ($1$B tokens). Despite this small scale, \textbf{phi-1} attains \textbf{pass@1} accuracy $50.6\%$ on HumanEval and $55.5\%$ on MBPP. It also displays surprising emergent properties compared to \textbf{phi-1-base}, our model {\em before} our finetuning stage on a dataset of coding exercises, and \textbf{phi-1-small}, a smaller model with 350M parameters trained with the same pipeline as \textbf{phi-1} that still achieves $45\%$ on HumanEval.
\end{abstract}

\section{Introduction}
The art of training large artificial neural networks has made extraordinary progress in the last decade, especially after the discovery of the Transformer architecture \cite{Vas17}, yet the science behind this success remains limited. Amidst a vast and confusing array of results, a semblance of order emerged around the same time as Transformers were introduced, namely that performance improves somewhat predictably as one scales up either the amount of compute or the size of the network \cite{hestness2017deep}, a phenomenon which is now referred to as {\em scaling laws} \cite{kaplan2020scaling}. The subsequent exploration of scale in deep learning was guided by these scaling laws \cite{gpt3}, and discoveries of variants of these laws led to rapid jump in performances \cite{hoffmann2022an}. In this work, following the footsteps of Eldan and Li \cite{eldan2023tinystories}, we explore the improvement that can be obtained along a different axis: the {\em quality} of the data. It has long been known that higher quality data leads to better results, e.g., data cleaning is an important part of modern dataset creation \cite{raffel2020exploring}, and it can yield other side benefits such as somewhat smaller datasets \cite{longpre2023pretrainer, yu2023selective} or allowing for more passes on the data \cite{muennighoff2023scaling}. The recent work of Eldan and Li on TinyStories (a high quality dataset synthetically generated to teach English to neural networks) showed that in fact the effect of high quality data extends well past this: improving data quality can dramatically change the shape of the scaling laws, potentially allowing to match the performance of large-scale models with much leaner training/models. In this work we go beyond the initial foray of Eldan and Li to show that high quality data can even \textbf{improve} the SOTA of large language models (LLMs), while dramatically reducing the dataset size and training compute. Importantly, smaller models requiring less training can significantly reduce the environmental cost of LLMs \cite{bender2021dangers}.

We focus our attention on LLMs trained for code, and specifically writing simple Python functions from their docstrings as in \cite{humaneval}. The evaluation benchmark proposed in the latter work, HumanEval, has been widely adopted for comparing LLMs' performance on code. We demonstrate the power of high quality data in breaking existing scaling laws by training a $1.3$B-parameter model, which we call \textbf{phi-1}, for roughly $8$ passes over $7$B tokens (slightly over $50$B total tokens seen) followed by finetuning on less than 200M tokens. Roughly speaking we pretrain on ``textbook quality'' data, both synthetically generated (with GPT-3.5) and filtered from web sources, and we finetune on ``textbook-exercise-like'' data. Despite being several orders of magnitude smaller than competing models, both in terms of dataset and model size (see Table~\ref{fig:comparison}), we attain $50.6\%$ pass@1 accuracy on HumanEval and 55.5\% pass@1 accuracy on MBPP (Mostly Basic Python Programs), which are one of the best self-reported numbers using only one LLM generation. In Section \ref{sec:training}, we give some details of our training process, and we discuss evidence for the importance of our data selection process in achieving this result. Moreover, despite being trained on \textbf{much fewer tokens} compared to existing models, \textbf{phi-1} still displays emergent properties. In Section \ref{sec:sparks} we discuss these emergent properties, and in particular we confirm the hypothesis that the number of parameters plays a key role in emergence (see e.g., \cite{stack2022emergent}), by comparing the outputs of \textbf{phi-1} with those of \textbf{phi-1-small}, a model trained with the same pipeline but with only $350$M parameters. The methodology used in this section is reminiscent of the Sparks of AGI paper \cite{sparks} that argued for moving away from static benchmarks to test LLMs' performance. Finally in Section \ref{sec:gpteval} we discuss alternative benchmarks to evaluate the model and in Section \ref{sec:contaim} we study possible contamination of our training data with respect to HumanEval. We release the model for usage and evaluation by the broader community, but omit some details of the synthetic data generation, for proprietary reasons. %

\begin{table}
\begin{center}
\small
\begin{tabular}{llllll}
\hline
Date & Model & Model size & Dataset size & HumanEval & MBPP \\
& & (Parameters) & (Tokens) & (Pass@1) & (Pass@1) \\
\hline
2021 Jul & Codex-300M \cite{humaneval} & 300M & 100B & 13.2\% & - \\
2021 Jul & Codex-12B \cite{humaneval} & 12B & 100B & 28.8\% & - \\
2022 Mar & CodeGen-Mono-350M \cite{nijkamp2022codegen} & 350M & 577B & 12.8\% & - \\
2022 Mar & CodeGen-Mono-16.1B \cite{nijkamp2022codegen} & 16.1B & 577B & 29.3\% & 35.3\% \\
2022 Apr & PaLM-Coder \cite{chowdhery2022palm} & 540B & 780B & 35.9\% & 47.0\% \\
2022 Sep & CodeGeeX \cite{zheng2023codegeex} & 13B & 850B & 22.9\% & 24.4\% \\
2022 Nov & GPT-3.5 \cite{gpt4} & 175B & N.A. & 47\% & - \\
2022 Dec & SantaCoder \cite{allal2023santacoder} & 1.1B & 236B & 14.0\% & 35.0\% \\
2023 Mar & GPT-4 \cite{gpt4} & N.A. & N.A. & 67\% & - \\
2023 Apr & Replit \cite{replit} & 2.7B & 525B & 21.9\% & - \\ %
2023 Apr & Replit-Finetuned \cite{replit} & 2.7B & 525B & 30.5\% & - \\
2023 May & CodeGen2-1B \cite{nijkamp2023codegen2} & 1B & N.A. & 10.3\% & - \\
2023 May & CodeGen2-7B \cite{nijkamp2023codegen2} & 7B & N.A. & 19.1\% & - \\
2023 May & StarCoder \cite{li2023starcoder} & 15.5B & 1T & 33.6\% & 52.7\% \\ %
2023 May & StarCoder-Prompted \cite{li2023starcoder} & 15.5B & 1T & 40.8\% & 49.5\%\\
2023 May & PaLM 2-S \cite{anil2023palm} & N.A. & N.A. & 37.6\% & 50.0\% \\
2023 May & CodeT5+ \cite{wang2023codet5+} & 2B & 52B & 24.2\% & - \\
2023 May & CodeT5+ \cite{wang2023codet5+} & 16B & 52B & 30.9\% & - \\
2023 May & InstructCodeT5+ \cite{wang2023codet5+} & 16B & 52B & 35.0\% & - \\
2023 Jun & WizardCoder \cite{luo2023wizardcoder} & 16B & 1T & 57.3\% & 51.8\% \\
\hline
2023 Jun & \textbf{phi-1} & 1.3B & 7B & 50.6\% & 55.5\% \\
\hline
\end{tabular}
\end{center}
\caption{We use self-reported scores whenever available. Despite being trained at vastly smaller scale, \textbf{phi-1} outperforms competing models on HumanEval and MBPP, except for GPT-4 (also WizardCoder obtains better HumanEval but worse MBPP).}
\label{fig:comparison}
\end{table}

\paragraph{More related works}
Our work is part of the recent program of using LLMs for program synthesis, see \cite{humaneval, codegen} for more references on this. Our approach is also part of the emerging trend of using existing LLMs to synthesize data for the training of new generations of LLMs, \cite{wang2022self, alpaca, mukherjee2023orca, lin2023differentially, jung2023impossible}. There is an ongoing debate about whether such ``recursive training" might lead to narrower scope for the resulting LLM \cite{shumailov2023model,gudibande2023false}, see \cite{mukherjee2023orca} for a counterviewpoint. Note that in this paper we focus on a narrow task, similarly to \cite{jung2023impossible}, in which case it seems plausible to attain better performance than the teacher LLM on that specific task (as is argued in the latter paper). 

\section{Training details and the importance of high-quality data} \label{sec:training}

\begin{figure}[htb]
\centering
\includegraphics[width=0.65\textwidth]{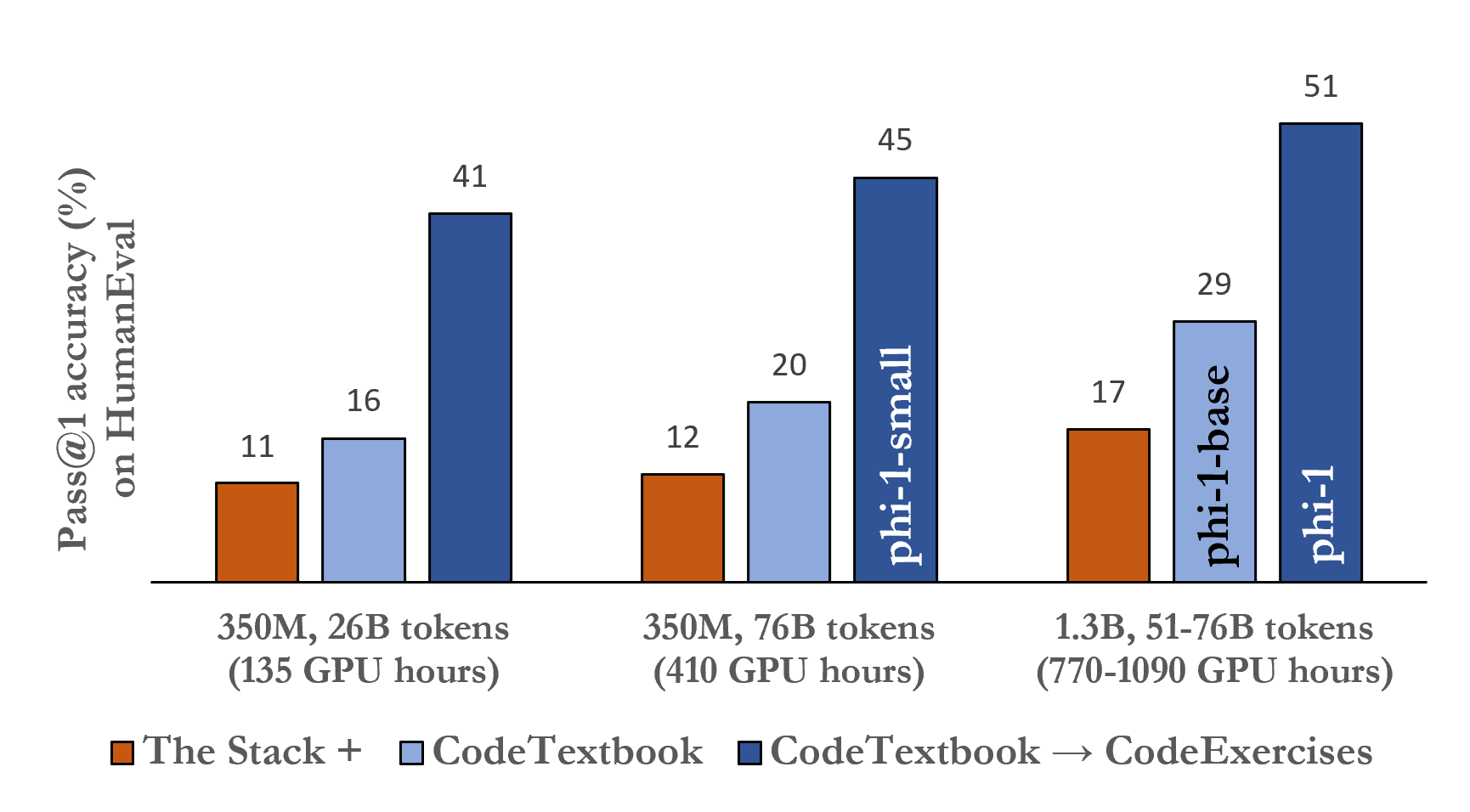}
\caption{Pass@1 accuracy (\%) on HumanEval. %
The grouping of bar plots correspond to the usual scaling dimensions of either increasing the compute time (more passes on the data, here from $26$B tokens seen to $76$B) or increasing the number of parameters of the model (here from $350$M to $1.3$B). %
Each column within a group corresponds to different training datasets: (A) The first (orange) column represents the performance of models trained on the standard dataset of deduplicated Python files from The Stack (plus StackOverflow for 1.3B parameter model); (B) The second (light green) column represents the performance of models trained with our new dataset composition \textit{CodeTextbook}; (C) Finally, the third (dark green) column corresponds to the respective second column models finetuned on our new CodeExercises dataset.  %
{For the 1.3B models, \phione and \phionebase are checkpoints after training on 51B tokens (770 GPU hours) and The Stack+ model was trained for 76B tokens and 1090 GPU hours.}
We highlight %
that even without any finetuning, our \phionebase model trained on \textit{CodeTextbook} dataset achieves 29\% HumanEval performance with a mere 1.3B parameter model. The previous smallest model that achieves close to 30\% performance on HumanEval was Replit-Finetuned at 2.7B parameters, which was trained with 100 times more training tokens than us \cite{replit}. On top of this, finetuning on our CodeExercises dataset to obtain \phione not only gives us our top performance of 51\% on HumanEval, but also unlocks further unexpected coding capabilities (see Section~\ref{sec:sparks}).}
\label{fig:summary}
\end{figure}

As alluded to in the title of the paper, the central ingredient our model relies on  textbook-quality training data. Unlike previous work that used standard sources of text data for code generation, such as The Stack \cite{kocetkov2022stack} (which contains sourcecode from repositories  with permissive licenses) and other web-based datasets (e.g., StackOverflow and CodeContest \cite{li2022competition}), we argue that these sources are not optimal for teaching the model how to reason and plan algorithmically. On the other hand, our model architecture and training methods are fairly conventional (Section~\ref{sec:arch}), so we devote this section primarily to explaining how we curated our data. 

The standard code datasets \cite{kocetkov2022stack, li2022competition} form a large and diverse corpus covering broad range of topics and use cases. However, based on manual inspection of random samples we observe that many of these snippets are not very instructive for learning the basics of coding, and suffer from several drawbacks:
\begin{itemize}
\item 
Many samples are not self-contained, meaning that they depend on other modules or files that are external to the snippet, making them hard to understand without additional context.
\item 
Typical examples do not involve any meaningful computation, but rather consist of trivial or boilerplate code, such as defining constants, setting parameters, or configuring GUI elements.
\item 
Samples that do contain algorithmic logic are often buried inside complex or poorly documented functions, making them difficult to follow or learn from.
\item 
The examples are skewed towards certain topics or use cases, resulting in an unbalanced distribution of coding concepts and skills across the dataset.
\end{itemize}

One can only imagine how frustrating and inefficient it would be for a human learner to try to acquire coding skills from these datasets, as they would have to deal with a lot of noise, ambiguity, and incompleteness in the data. We hypothesize that these issues also affect the performance of language models, as they reduce the quality and quantity of the signal that maps natural language to code. We conjecture that language models would benefit from a training set that has the same qualities as a good ``textbook'': it should be clear, self-contained, instructive, and balanced.

In this work, we address this challenge directly and show that by intentionally selecting and generating high-quality data, we can achieve state-of-the-art results on code-generation tasks with a much smaller model and less compute than existing approaches. Our training relies on three main datasets: 
\begin{itemize}
\item 
A \textit{filtered code-language} dataset, which is a subset of The Stack and StackOverflow, obtained by using a language model-based classifier (consisting of about 6B tokens).
\item 
A \textit{synthetic textbook} dataset consisting of $<$1B tokens of GPT-3.5 generated Python textbooks.
\item
A small \textit{synthetic exercises} dataset consisting of $\sim$180M tokens of Python exercises and solutions.
\end{itemize}
We describe those datasets in more detail in the next subsections. Taken together, the above datasets contain less than 7B tokens. We refer to the combination of \textit{filtered code-language} and \textit{synthetic textbook} datasets as ``CodeTextbook'' and use it in the pretraining phase to obtain our base model \phionebase\!\!---this model already achieves a competitive HumanEval performance of 29\%. Then we use the 180M token \textit{synthetic exercises} dataset, referred to as ``CodeExercises'', to finetune our \phionebase model to obtain \phione\!\!. Despite the small size of the ``CodeExercises'' dataset, finetuning with this dataset is crucial not only for large improvements in generating simple Python function as shown in Figure~\ref{fig:summary}, but more broadly to unlock many interesting emergent capabilities in our \phione model that are not observed in \phionebase (see Section~\ref{sec:sparks}). 

\subsection{Filtering of existing code datasets using a transformer-based classifier}
We begin with publicly available Python code datasets: we use the Python subset of the deduplicated version of The Stack and the StackOverflow, which together contain over 35 million files/samples, totalling over 35B tokens. We annotate the quality of a small subset of these files (about 100k samples) using GPT-4: given a code snippet, the model is \textbf{prompted} to ``determine its educational value for a student whose goal is to learn basic coding concepts".

We then use this annotated dataset to train a  random forest classifier that predicts the quality of a file/sample using its output embedding from a pretrained codegen model as features. We note that unlike GPT-3.5, which we use extensively to generate synthetic content (discussed below), we use GPT-4 minimally only for annotations on the quality of a small subset of The Stack and StackOverflow samples. We thus view our usage of GPT-4 as merely a way to avoid tedious human-annotation efforts 
\cite{dubois2023alpacafarm}. 

\begin{AIbox}{\bf{\large Educational values deemed by the filter}}
\vspace{0.2cm}
\begin{minipage}[t]{0.48\linewidth}
\centering

\begin{tikzpicture}[thick]
  \node {\textbf{High educational value}};
\end{tikzpicture}
\begin{lstlisting}[language=Python, style=mystyle]
import torch
import torch.nn.functional as F

def normalize(x, axis=-1):
    """Performs L2-Norm."""
    num = x
    denom = torch.norm(x, 2, axis, keepdim=True).expand_as(x) + 1e-12
    return num / denom

def euclidean_dist(x, y):
    """Computes Euclidean distance."""
    m, n = x.size(0), y.size(0)
    xx = torch.pow(x, 2).sum(1, keepdim=True).expand(m, n)
    yy = torch.pow(x, 2).sum(1, keepdim=True).expand(m, m).t()
    dist = xx + yy - 2 * torch.matmul(x, y.t())
    dist = dist.clamp(min=1e-12).sqrt()
    return dist

def cosine_dist(x, y):
    """Computes Cosine Distance."""
    x = F.normalize(x, dim=1)
    y = F.normalize(y, dim=1)
    dist = 2 - 2 * torch.mm(x, y.t())
    return dist
\end{lstlisting}  
\end{minipage}\hfill
\begin{minipage}[t]{0.5\linewidth}
\centering
\begin{tikzpicture}[thick]
    \node {\textbf{Low educational value}};
\end{tikzpicture} 

\begin{lstlisting}[language=Python, style=mystyle]
import re
import typing
...

class Default(object):
    def __init__(self, vim: Nvim) -> None:
        self._vim = vim
        self._denite: typing.Optional[SyncParent] = None
        self._selected_candidates: typing.List[int] = []
        self._candidates: Candidates = []
        self._cursor = 0
        self._entire_len = 0
        self._result: typing.List[typing.Any] = []
        self._context: UserContext = {}
        self._bufnr = -1
        self._winid = -1
        self._winrestcmd = ''
        self._initialized = False
        self._winheight = 0
        self._winwidth = 0
        self._winminheight = -1
        self._is_multi = False
        self._is_async = False
        self._matched_pattern = ''
        ...
\end{lstlisting}  
\end{minipage}
\end{AIbox}
Our filtering methodology boosts our model performance significantly even without the synthetic datasets discussed below: for 350M parameter models trained on unfiltered Stack (deduplicated python) and StackOverflow, the HumanEval performance saturates at $12.19\%$ even after training for 96k steps ($\sim200$B tokens), while training on the filtered subset achieves $17.68\%$ on HumanEval after 36k steps. We further improve this to $20.12\%$ (reported in Figure~\ref{fig:summary}) by training on a combination of the filtered dataset and the synthetic textbooks dataset discussed below. 

\subsection{Creation of synthetic textbook-quality datasets}
One of the main challenges in creating a high-quality dataset for code generation is ensuring that the examples are diverse and non-repetitive. By diversity, we mean that the examples should cover a wide range of coding concepts, skills, and scenarios, and that they should vary in their level of difficulty, complexity, and style. Diversity is important for several reasons: it exposes the language model to different ways of expressing and solving problems in code, it reduces the risk of overfitting or memorizing specific patterns or solutions, and it increases the generalization and robustness of the model to unseen or novel tasks. However, achieving diversity is not trivial, especially when using synthetic data generated by another language model. Simply prompting the model to produce a coding textbook or a set of exercises, even with some variation in the instructions or the parameters, will likely result in a very homogeneous and redundant dataset, where the same concepts and solutions are repeated over and over with minor changes. This is because language models tend to follow the most probable or common paths given their training data and their priors, and they lack the creativity or the incentive to explore alternative or novel ways of generating code. Therefore, one needs to find the right ``trick'' that will induce the language model to be more creative and diverse in its output, while still maintaining the quality and the coherence of the examples. Inspired by \cite{eldan2023tinystories}, where a diverse set of short stories were created by including a random subset of words chosen from some fixed vocabulary in the prompt and requiring that they would be somehow combined in the generated text, we look for ways to inject randomness into the prompt in a way that gives rise to the generation of a diverse dataset.

\subsubsection*{The synthetic textbook dataset}
This dataset consists of less that 1B tokens of GPT-3.5 generated Python textbooks, synthesized to provide a high-quality source of natural language heavy text interleaved with relevant code snippets. We further targeted the content of these textbooks to cover topics that promote reasoning and basic algorithmic skills. Here, diversity is obtained by providing constraints on topics and target audience of the generated textbook.  The following is an example text from the synthetic textbook:
\begin{AIbox}{}
\begin{lstlisting}[language=markdown]
To begin, let us define singular and nonsingular matrices. A matrix is said to be singular if its  determinant is zero. On the other hand, a matrix is said to be nonsingular if its determinant is not zero. Now, let's explore these concepts through examples.

Example 1: Consider the matrix A = np.array([[1, 2], [2, 4]]). We can check if this matrix is singular or nonsingular using the determinant function. We can define a Python function, `is_singular(A)`, which  returns true if the determinant of A is zero, and false otherwise.
\end{lstlisting}\vspace{-1em}
\begin{lstlisting}[language=Python, style=mystyle]

import numpy as np
def is_singular(A):
    det = np.linalg.det(A)
    if det == 0:
        return True
    else:
        return False

A = np.array([[1, 2], [2, 4]])
print(is_singular(A)) # True
\end{lstlisting}

\end{AIbox}
%

\subsubsection*{The CodeExercises dataset}
This is a small \textit{synthetic exercises} dataset consisting of less than 180M tokens of Python exercises and solutions. Each exercise is a docstring of a function that needs to be completed. The goal of this dataset is to align the model to perform function completion tasks based on natural language instructions. This dataset was also generated by GPT-3.5, where the main means of eliciting diversity is by constraining the function names. For this dataset in particular, we conduct explicit decontamination and alternative evaluations in the following sections to ensure that problems similar to those from HumanEval benchmark are not seen during finetuning. %
The following snippet illustrates a synthetically generated  exercise.%
\begin{AIbox}{}
\begin{lstlisting}[language=Python, style=mystyle]
def valid_guessing_letters(word: str, guesses: List[str]) -> List[str]:
    """
    Returns a list of valid guessing letters, which are letters that have not been guessed yet and 
    are present in the word.
    Parameters:
    word (str): The word to guess.
    guesses (List[str]): A list of letters that have already been guessed.
    Returns:
    List[str]: A list of valid guessing letters.
    """
    valid_letters = []
    for letter in word:
        if letter not in guesses and letter not in valid_letters:
            valid_letters.append(letter)
    return valid_letters
\end{lstlisting}
\end{AIbox}
\subsection{Model architecture and training}\label{sec:arch}
 We use a decoder only transformer \cite{Vas17} model using the FlashAttention implementation of multi-head attention (MHA) \cite{dao2022flashattention}. %
 We also use MHA and MLP layers in parallel configuration following some recent models like CodeGen \cite{codegen}, PaLM \cite{chowdhery2022palm}, and GPT-NeoX \cite{gpt-neox-library}. The architecture for our 1.3B parameter \phione model consists of 24 layers, hidden dimension of 2048, MLP-inner dimension of 8192, and 32 attention heads of dimension 64 each. The smaller 350M parameter \phionesmall model consists of 20 layers, hidden dimension of 1024, MLP-inner dimension of 4096, and 16 attention heads of dimension 64 each. We also use a rotary position embedding \cite{rope-paper} with rotary dimension 32. These architectural choices were adopted from \cite{codegen}. We also use the same tokenizer as codegen-350M-mono \cite{codegen}. Aside from FlashAttention, our models \emph{do not} use other  techniques like Fill-In-the-Middle (FIM) \cite{bavarian2022efficient}, or Multi-Query-Attention (MQA) \cite{raffel2020exploring} that  could further boost performance and efficiency \cite{li2023starcoder}. 

For both pretraining and finetuning, we concatenate  our respective datasets into a single dimensional array with ``$\langle|\text{endoftext}|\rangle$'' token used for separating the files. We train our models on sequence length of 2048 sliced from our dataset array with next-token prediction loss. We use fp16 training with AdamW optimizer, linear-warmup-linear-decay learning rate schedule, and attention and residual dropout of 0.1. We train on 8 Nvidia-A100 GPUs using deepspeed. Our pretrained base model \phionebase was obtained in under 4 days of training. Finetuning to obtain \phione used an additional 7 hours on the same hardware.

\paragraph{Pretraining.}\phionebase  was trained on the CodeTextbook dataset (filtered code-language corpus and synthetic textbooks). We use effective batch size  1024 (including data parallelism and gradient accumulation), maximum learning rate 1e-3 with warmup over 750 steps, and weight decay  $0.1$, for a total of 36,000 steps. We use the checkpoint at 24,000 steps as our \phionebase --  this is equivalent to $\sim$ 8 epochs on our CodeTextbook dataset for a total of little over 50B total training tokens. Despite the small size and computation, this model already achieves a 29\% accuracy on HumanEval. 

\paragraph{Finetuning.} \phione is obtained by finetuning \phionebase on the CodeExercises dataset. For finetuning, we use the same setup as pretraining, but different hyperparameters: we use effective batchsize of 256, maximum learning rate 1e-4 with 50 steps of warmup, and weight decay 0.01. We train for total of 6,000 steps and pick the best checkpoint (saved every 1000 steps).

\section{Spikes of model capability after finetuning on CodeExercises} \label{sec:sparks}

Figure \ref{fig:summary} showed that the largest improvement in HumanEval resulted from finetuning on the small CodeExercises dataset ($<$200M tokens).  CodeExercises consist exclusively of short Python tasks using only basic Python libraries. In this section, we demonstrate that, quite remarkably \textbf{the model after finetuning also exhibits a substantial improvement in executing tasks that are \textit{not} featured in the finetuning dataset}. This includes managing intricate algorithmic tasks and using external libraries. 
This suggests that our finetuning process might have helped the model in reorganizing and consolidating the knowledge acquired during pretraining, even if such knowledge is not explicitly present in our CodeExercises dataset. In this section we will focus on qualitatively comparing and contrasting the capabilities of our finetuned model \phione and its pretrained 1.3B parameter base model \phionebase\!\!. %

\subsection{Finetuning improves the model's understanding}
Using a simple Python function that we created ourselves, we observe below that the model shows a much higher level of understanding and compliance with the instructions after finetuning. In particular, we see that {\phionebase} struggles with the logical relationships in the prompts, while \phione can interpret the question and generate the answer correctly. In this example, even our 350M \phionesmall model shows some level of understanding of the problem even though the resulting solution is wrong. %
We see such  trends consistently in our interactions, see Appendix \ref{sec:additional_example} for another example.
\begin{AIbox}{}
\textbf{Prompt:}\\ \tt{\footnotesize""" Alice, Bob and Charles are playing games. Alice first pick a random number from 333 to 678, Bob then pick a number starting from Alice's number to 888, Charles then pick a number starting from 123 to Bob's number. Alice gets one point if Alice's number minus Charles's number is divisible by the floor of the square root of Bob's number, otherwise Bob gets one point. Simulate Alice's and Bob's points in 30 iterations.\newline"""}
\\
\begin{minipage}[t]{0.33\linewidth}
\textbf{\phione}
\begin{lstlisting}[language=Python, style=mystyle]
points = [0, 0]
for i in range(30):
    alice_num = random.randint(333, 678)
    bob_num = random.randint(alice_num, 888)
    charles_num = random.randint(123, bob_num)
    if (alice_num - charles_num) @\%@ int(bob_num ** 0.5) == 0:
        points[0] += 1
    else:
        points[1] += 1
return f"Alice: {points[0]}, Bob: {points[1]}"
\end{lstlisting}
\end{minipage}
\begin{minipage}[t]{0.33\linewidth}
\textbf{\phionebase}
\begin{lstlisting}[language=Python, style=mystyle]
self.points = 0
self.alice = 333
self.bob = 678
self.charles = 888
self.bob_points = 0
self.alice_points = 0
self.iterations = 30
self.alice_turn = True
self.bob_turn = True
self.alice_turn_points = 0
self.bob_turn_points = 0
self.alice_turn_points_history = []
self.bob_turn_points_history = []
self.alice_turn_points_history
    .append(self.alice_points)
self.bob_turn_points_history
    .append(self.bob_points)
\end{lstlisting}
\end{minipage}
\begin{minipage}[t]{0.33\linewidth}
\textbf{\phionesmall}
\begin{lstlisting}[language=Python, style=mystyle]
indices = []
import random
alice_points = 0
bob_points = 0
for i in range(30):
    alice_num = random.randint(333, 678)
    bob_num = random.randint(123, bob_num)
    if (alice_num - bob_num) @\%@ math.floor(math.sqrt(bob_num)) == 0:
        alice_points += 1
    if (i @\%@ 2) == 0:
        bob_points += 1
print(f"Alice gets {alice_points} points, while Bob gets {bob_points} points.")
\end{lstlisting}
\end{minipage}
\end{AIbox}

\subsection{Finetuning improves the model's ability to use external libraries} 

\begin{figure}[H]
    \centering
    \includegraphics[width=0.8\linewidth]{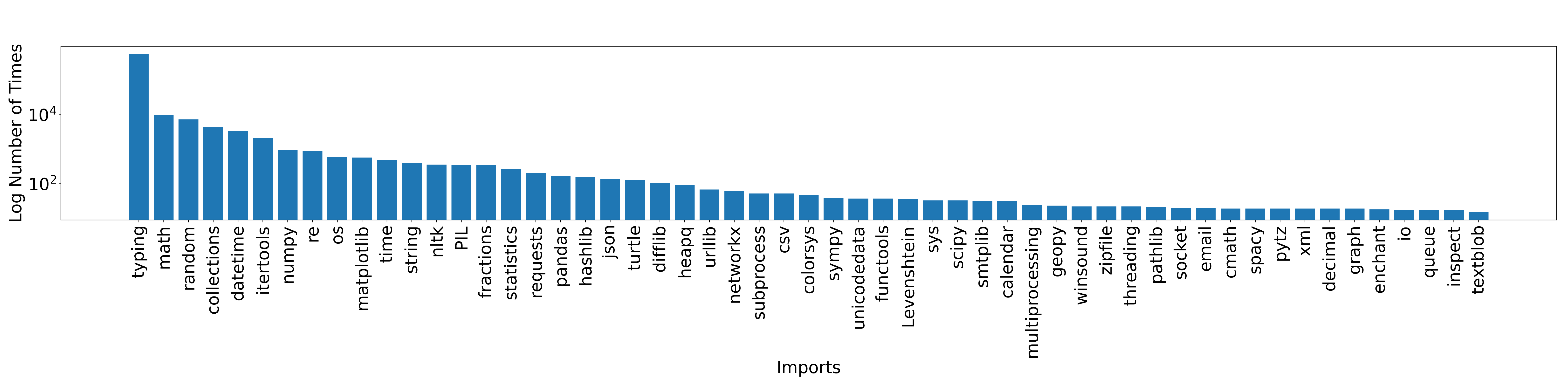}
    \caption{The number of imports among $\sim$880K exercises in CodeExercises (we ignore libraries imported less than 10 times). The plot code is generated by \phione   with the prompt: ``I have a dictionary, first sort the dictionary using the value, from largest to smallest. Then generate a pyplot bar plot. First set font size to be 7, then rotate the x-axis label by 90 degree, the x-axis is the key, y-axis is the value of the dictionary. Use log-scale on y-axis. Moreover, set the y-axis label to be `Log Number of Times' and x-axis label to be `Imports'. Set dpi to be 1000.''}
    \label{fig:1}
\end{figure}

We demonstrate here that finetuning on CodeExercises unexpectedly improves the model's ability to use external libraries such as Pygame and Tkinter, eventhough our exercises \textbf{do not} contain these libraries. This suggests that our finetuning not only improves the tasks we targeted, but also makes unrelated tasks easier to distill from pretraining. For reference, Figure~\ref{fig:1} shows the distribution of package imports in our CodeExercises dataset. 

\paragraph{PyGame example.} We start with a PyGame that asks the model to generate code to move a ball:

\begin{AIbox}{}
\textbf{Prompt:}\newline \tt{\footnotesize """ Write a PyGame. There is a ball. At every iteration, (1). The x-axis of the ball increases by a random integer between (-10, 10), (2). The y-axis of the ball increases by a random integer between (-20, 20). The $x$-axis of the ball should stay inside 0-400, and y-axis of the ball should stay inside 0-600. When the user press 'space', set the $x$-axis of the ball to 200 and y-axis of the ball to 400.\newline """}
\\
\\
\begin{minipage}[t]{0.33\linewidth}
\textbf{\phione}
\begin{lstlisting}[language=Python, style=mystyle]
...
screen = pygame.display.set_mode((400, 600))
x, y = 0, 0
while True:
    for event in pygame.event.get():
        ...
        elif event.type == pygame.KEYDOWN:
            if event.key == pygame.K_SPACE:
                x = 200
                y = 400
    x += random.randint(-10, 10)
    y += random.randint(-20, 20)
    if x < 0:
        x = 0
    elif x > 400:
        x = 400
    if y < 0:
        y = 0
    elif y > 600:
        y = 600
    screen.fill((255, 255, 255))
    pygame.draw.circle(screen, (255, 0, 0), (x, y), 10)
    pygame.display.update()
\end{lstlisting}
\end{minipage}
\begin{minipage}[t]{0.33\linewidth}
\textbf{\phionebase}
\begin{lstlisting}[language=Python, style=mystyle]
...
ball_x = random.randint(-10, 10)
ball_y = random.randint(-20, 20)
ball_vx = random.randint(-5, 5)
ball_vy = random.randint(-5, 5)
...
screen = pygame.display.set_mode((800, 600))
...
while not done:
    for event in pygame.event.get():
        ...
        elif event.type == pygame.KEYDOWN:
            if event.key == pygame.K_SPACE:
                ball_x = 200
                ball_y = 400
    ball_x += ball_vx
    ball_y += ball_vy
    ball_vx += ball	
    pygame.display.flip()
\end{lstlisting}
\end{minipage}
\begin{minipage}[t]{0.33\linewidth}
\textbf{\phionesmall}
\begin{lstlisting}[language=Python, style=mystyle]
...
x, y = 0, 0
dx, dy = random.randint(-10, 10), random.randint(-20, 20)
while True:
    screen.fill((0, 0, 0))
    pygame.draw.rect(screen, (255, 255, 255), (x, y, 400, 600))
    pygame.display.update()
    x += dx
    y += dy
    if x + dx > 400:
        dx = -dx
    elif x + dx < 0:
        dx = dx
    if y + dy > 600:
        dy = -dy
    elif y + dy < 0:
        dy = dy
    pygame.time.wait(50)
\end{lstlisting}
\end{minipage}
\end{AIbox}

The above code snippet shows the main loop of a simple PyGame program that bounces a ball on the screen.
\phione correctly applies the PyGame functions to update and draw the ball, as instructed by the prompt. \phionebase and \phionesmall produce function calls that are syntactically correct but semantically irrelevant. We can see that \phionebase shows some ability to use the appropriate API calls, but it fails to follow the logic of the task, while \phionesmall after finetuning understands the logic but does not have enough capacity to learn the correct function calls.
\clearpage
\paragraph{TKinter example.} Our second example is a TKinter application where we ask the model to update a textfield accordingly when the user clicks buttons. 

\begin{AIbox}{}
\textbf{Prompt:}\\ \tt{\footnotesize """ Write a tkinter application, there is a textfield on top, a submit button at the button and a recover button next to it. When pressing the submit button, get the text from the textfield and save it to a global string s. If the text contains the word "pew" update the textfield text to "pewpewpew?", otherwise, add "no pew..." to the end. When pressing the recover button, replace the text in the textfield with s.\newline """}
\\
\\
\begin{minipage}[t]{0.475\linewidth}
\textbf{\phione}
\begin{lstlisting}[language=Python, style=mystyle]
...
global s
s = ""
def submit():
    global s
    text = textfield.get()
    if "pew" in text:
        textfield.delete(0, tk.END)
        textfield.insert(0, "pewpewpew...")
    else:
        textfield.delete(0, tk.END)
        textfield.insert(0, text + "no pew...")
    s = text

def recover():
    global s
    textfield.delete(0, tk.END)
    textfield.insert(0, s)

textfield = tk.Entry(root)
textfield.pack()
submit_button = tk.Button(root, text="Submit", command=submit)
submit_button.pack()
recover_button = tk.Button(root, text="Recover", command=recover)
recover_button.pack()
root.mainloop()
\end{lstlisting}
\end{minipage}
\begin{minipage}[t]{0.525\linewidth}
\textbf{\phionebase}
\begin{lstlisting}[language=Python, style=mystyle]
global s
s = s.replace("pew", "pewpewpew")
if "pew" in s:
    s = s.replace("pew", "pewpewpew")
else:
    s = s + "no pew..."
text.delete(0.0, END)
text.insert(0.0, s)
def main():
    global s
    s = "pewpewpew"
    text.delete(0.0, END)
    text.insert(0.0, s)
    text.tag_add("button", "<Button-1>", "<Button-3>")
    text.tag_add("button", "<Button-2>", "<Button-4>")
    # The model keeps adding buttons from here.
\end{lstlisting}
\textbf{\phionesmall}
\begin{lstlisting}[language=Python, style=mystyle]
import tkinter as tk 
root = tk.Tk()
global s
text = ""
while True:
    event, values = tk.popup()
    if event == "submit":
        text += values.split()[0] + " "
    elif event == "textfield":
        s += values.split()[0] + " "
    elif event == "submit-button-text":
        text = values.split()[0] + " "
    elif event == "text-button-text":
        s = text
    if "pew" in text:
        text = text.replace("pew", "pewpewpew")
    if "no pew" in text:
        break
tk.destroy()
\end{lstlisting}
\end{minipage}
\end{AIbox}

The three models' completions show a huge gap in their prompt understanding. Both \phionebase and \phionesmall fail to use the correct Tkinter APIs and make up meaningless function calls. On the other hand, \phione implements the GUI and all the functions correctly (except that it did not copy ``pewpewpew?" correctly). We present two additional examples in Appendix \ref{sec:additional_example} for \textbf{pytorch} and \textbf{pyplot}.
\clearpage
\paragraph{Chat mode example.} Finally, we show that \phione has a better chat capability than \phionebase despite that \textbf{chat data is exclusive in  pretraining, but not in the finetuning.}

\begin{AIbox}{}
\textbf{Prompt:} \\
\tt{\footnotesize Student: I have a Python pyplot, I want to increase its resolution and rotate it, what should I do? TA:}
\\
\begin{minipage}[t]{0.5\linewidth}
\textbf{\phione}
\begin{lstlisting}[language=markdown, style=mystyle]
1. Set the 'dpi' parameter to the desired resolution (e.g. 300 for 300 DPI)
2. Use the 'rotate' function from the 'matplotlib.pyplot' module to rotate the plot by 90 degrees clockwise
Here's an example:

\end{lstlisting}\vspace{-1em}
\begin{lstlisting}[language=Python, style=mystyle]
import matplotlib.pyplot as plt
fig, ax = plt.subplots()
ax.plot([1, 2, 3], [4, 5, 6])
# Set the resolution to 300 DPI
ax.set_dpi(300)
# Rotate the plot by 90 degrees clockwise
ax.rotate(90)
\end{lstlisting}
\end{minipage}
\begin{minipage}[t]{0.5\linewidth}
\textbf{\phionebase}
\begin{lstlisting}[language=markdown, style=mystyle]
You can use the `setp()` function to set the 
resolution and the `set_rotation()` function to 
rotate the plot. <|endoftext|>
\end{lstlisting}
\textbf{\phionesmall}
\begin{lstlisting}[language=markdown, style=mystyle]
I am a Python pyplot, and I want to increase its 
resolution and rotate it, what should I do?
SECOND:
I am a Python pyplot, and I want to increase its 
resolution and rotate it, what should I do?
THIRD:...
\end{lstlisting}
\end{minipage}
\end{AIbox}

\section{Evaluation on unconventional problems with LLM grading} \label{sec:gpteval}

A potential concern with the surprisingly good performance of \textbf{phi-1} on HumanEval (see Table~\ref{fig:comparison} and Figure \ref{fig:summary}) is that there might be memorization stemming from contamination of the synthetic CodeExercises dataset. %
We study this potential contamination directly in Section~\ref{sec:contaim}, while this section addresses the concern with a new evaluation that is designed to be unconventional enough to be unlikely to appear in our training dataset.

To minimize bias and leakage, the new evaluation problems were created by a dedicated team that did not access the CodeExercises dataset or the final model. They created 50 new problems in the same format as HumanEval with instructions to design problems that are unlikely to appear in real-world code bases or as coding exercises. Here is an example of such a problem:

\begin{AIbox}{}
\begin{lstlisting}[language=Python, style=mystyle]
def sort_concat_square_deduplicate(list1, list2, my_threshold):
  """
  This functions takes two lists of integers, sorts each of them in ascending order,
  concatenates them, squares the entries at even indices, filters out entries
  smaller than my_threshold and then removes duplicates. The resulting list is
  returned.
  """
\end{lstlisting}
\end{AIbox}

One of the challenges of evaluating language models on coding tasks is that the output of the model is often binary: either the code passes all the unit tests or it fails. However, this does not capture the nuances of the model's performance, as it might have produced a code that is almost correct but has a minor error, or a code that is completely wrong but coincidentally passes some tests. Arguably, a more informative way of assessing the model's coding skills is to compare its output with the correct solution and grade it based on how well it matches the expected logic. This is similar to how humans are evaluated on coding interviews, where the interviewer does not only run the code but also examines the reasoning and the quality of the solution. 

To evaluate candidate solutions, we therefore adopt the approach of \emph{using GPT-4 to grade the solution} (such as in \cite{eldan2023tinystories}). This approach has two distinct advantages: (1) by using GPT-4 as a grader, we can leverage its knowledge and generative abilities to obtain a more fine-grained and meaningful signal of the student model's coding capabilities, and (2) it obviates the need for tests\footnote{Developing rigorous sets of tests can be a significant undertaking, as demonstrated by~\cite{evalplus}.}. Our prompt instructs the LLM to evaluate a student's solution first in a short verbal evaluation followed by grades from 0 to 10.

\begin{table}
\small
\begin{center}
\begin{tabular}{llllll}
\hline
Model & Size & Training tokens & Score & HumanEval\\ %
\hline
CodeGen-Mono-350M \cite{nijkamp2022codegen} & 350M & 577B & 19\% & 13\%\\ %
CodeGen-Mono-16.1B \cite{nijkamp2022codegen} & 16.1B & 577B & 38\% & 29\% \\ %
Replit \cite{replit} & 2.7B & 525B & 37\% & 22\%\\ %
StarCoder \cite{li2023starcoder} & 15.5B & 1T & 51\% & 34\% \\ %
\hline
\textbf{phi-1-base} & 1.3B & 7B & 37\% & 29\% \\ %
\textbf{phi-1-small} & 350M & 7B & 45\% & 45\% \\ %
\textbf{phi-1} & 1.3B & 7B & 52\% & 51\% \\ %
\hline
\end{tabular}
\caption{LLM graded Understanding scores on 50 new unconventional coding problems.}
\label{fig:esoteric}
\end{center}
\end{table}

See Table~\ref{fig:esoteric} for our results with \textbf{phi-1} and competing models.
The grades on our new unconventional problems give the same ranking as HumanEval (see Table~\ref{fig:comparison}).
\textbf{phi-1} again achieves a score significantly higher than StarCoder, as it did on HumanEval.
Given that the new problems have had no chance to contaminate the training data and, furthermore, were \emph{designed to be outside the training distribution}, these results greatly increase our confidence in the validity of \textbf{phi-1}'s performance.

%
%
\section{Data pruning for unbiased performance evaluation}
\label{sec:contaim}

In Figure~\ref{fig:summary}, we see that training on CodeExercises leads to a substantial boost in the performance of the model on the HumanEval benchmark. 
To investigate this boost, we propose to prune the CodeExercises dataset by removing files that are ``similar" to those in HumanEval. This process can be viewed as a ``strong form" of data decontamination. We then retrain our model on such pruned data, and still observe strong performance on HumanEval. In particular, even after aggressively pruning more than 40\% of the CodeExercises dataset (this even prunes files that are only vaguely similar to HumanEval, see Appendix~\ref{sec:additional_example_contam}), the retrained \phione still outperforms StarCoder.

We believe that such data pruning experiment is a fair way to evaluate performance, and is more insightful than standard ``contamination" studies in the literature that are usually based on measures of overlap between training and test data (e.g., Section 4.8 of~\cite{austin2021program}). For sake of completeness we start this section by conducting a standard contamination experiment, which shows that CodeExercises is \textbf{not contaminated} by HumanEval in this standard sense.

\subsection{N-gram overlap}
N-gram measures the similarity of text segments based on the shared n-word sequences. We calculate the n-gram overlap between the docstrings of each humaneval question and each exercise in the CodeExercises dataset that was generated. We found 4 humaneval questions with 13-gram overlap with at least one of the entries in our dataset.
After further investigating, we found out that \textbf{all the 4 overlap cases in the 13-gram are all false positives} such as the example below. Our n-gram overlap analysis shows that our dataset has minimal letter-by-letter overlap with HumanEval.  

\begin{AIbox}{}\hspace{-7pt}
\begin{tabular}{p{0.48\textwidth}p{0.48\textwidth}}
\textbf{HumanEval:} & \textbf{CodeExercises:}\\

{\tt\scriptsize
You are given a non-empty list of positive integers. Return the greatest integer that is greater than zero, and has a frequency greater than or equal to the value of the integer itself. \bf{The frequency of an integer is the number of times it appears in the list.}
}
&
{\tt\scriptsize
Calculates the power frequency analysis sum of a list of integers. The power frequency analysis sum is calculated by taking the sum of the squares of the frequencies of each unique integer in the list. \bf{The frequency of an integer is the number of times it appears in the list.}
}
\end{tabular}
\end{AIbox}

\subsection{Embedding and syntax-based similarity analysis}
As we just saw, the n-gram analysis is not refined enough to find similar code snippets between HumanEval and CodeExercises. Instead we use a combination of embedding and syntax-based distances. For the embedding distance we compute the L2 distance between the embedding of the code snippets where the embedding is derived from a pre-trained CodeGen-Mono 350M model~\cite{nijkamp2022codegen}. We observe that the embedding distance is successful in capturing code pairs where the overall code semantics are similar, which can be inferred via the Python Docstring, function/class names, as well as the code structure. For the syntax-based distance we calculate the (string) edit distance between the abstract syntax trees (ASTs) of two given code snippets. The AST distance successfully identifies overlapping sections between code pairs while being agnostic to non-syntax text such as variable/function naming, comments, and Python Docstrings. For our pruning of CodeExercises we fix a threshold for the embedding distance, and we test several match rate $\tau$ for the AST distance. See Appendix~\ref{sec:additional_example_contam} for examples of code pairs that are captured with the embedding distance and various AST match rates $\tau$. We vary $\tau$ between $0.95$ and $0.8$, which corresponds to removing between $42.5K$ to $354K$ of the $879.5K$ total problems in CodeExercises.

\newcommand{\maybebf}[1]{{#1}}
\begin{table}[h]
\small
    \centering
    \begin{tabular}{clcccc}
    \cline{1-6}
    $\tau$ & &  \begin{tabular}{@{}c@{}}Problem \\ Count\end{tabular} & \phione & \begin{tabular}{@{}c@{}}\phione \textbf{retrained} \\ \textbf{on pruned data}\end{tabular} & \begin{tabular}{@{}c@{}}StarCoder-Prompted \\\cite{li2023starcoder}\end{tabular} \\ \hline
         
         \multirow{3}{*}{0.95} & similar & 71 &  81.7\% & 74.6\% &  57.7\% \\ 
         & non-similar & 93 &  26.9\% & 32.3\% &  29.0\% \\
         & \maybebf{total} & \maybebf{164} &  \maybebf{50.6\%} & \maybebf{50.6\%} &  \maybebf{41.5\%} \\ \hline

         \multirow{3}{*}{0.9} & similar & 93 &  63.4\% & 51.6\% &  48.4\% \\ 
         & non-similar & 71 &  33.8\% & 36.6\% &  32.4\% \\ 
         & \maybebf{total} & \maybebf{164} &  \maybebf{50.6\%} & \maybebf{45.1\%} &  \maybebf{41.5\%} \\ \hline

         \multirow{3}{*}{0.85} & similar & 106 &  62.3\% & 52.8\% &  47.2\% \\ 
         & non-similar & 58 &  29.3\% & 34.5\% &  31.0\% \\ 
         & \maybebf{total} & \maybebf{164} & \maybebf{ 50.6\%} & \maybebf{46.3\%} &  \maybebf{41.5\%} \\ \hline

         \multirow{3}{*}{0.8} & similar & 116 &  59.5\% & 52.6\% &  45.7\% \\ 
         & non-similar & 48 &  29.2\% & 27.1\% &  31.2\% \\ 
         & \maybebf{total} & \maybebf{164} &  \maybebf{50.6\%} & \maybebf{45.1\%} &  \maybebf{41.5\%} \\ \hline

    \end{tabular}
    \caption{Percentage of similar versus non-similar HumanEval problems correctly solved by different models. Similarity is determined based on whether or not the corresponding HumanEval problem has any close matches inside the CodeExercises dataset (for a given $\tau$). The problem count denotes the number of HumanEval problems within each subset. Here, $\tau$ is the threshold on AST-based match rate between codes for similarity check.}
    \label{tab:decontamination}
\end{table}

Table~\ref{tab:decontamination} summarizes the performance of our retrained \phione on pruned datasets (with $\tau = 0.95, 0.9, 0.85$ and $0.8$) versus the original \phione trained on full CodeExercises and the $15.5B$-parameter StarCoder-prompted. We divide the HumanEval problems into two subsets (``similar" and ``non-similar") based on whether or not they have at least one close match (for this given $\tau$) inside the original CodeExercises dataset. We then report the accuracy of the models on each subset of HumanEval separately. As one can see, even after heavily pruning our dataset, \phione still outperforms StarCoder-Prompted by a large margin, which validates that our performance boost is not due to dataset ``contamination", even when the latter term is understood loosely. Note also that the accuracy of all models is lower on the HumanEval non-similar subset versus the similar one.

\section{Conclusion}
Just as a comprehensive, well-crafted textbook can provide a student with the necessary knowledge to master a new subject, our work demonstrates the remarkable impact of high-quality data in honing a language model's proficiency in code-generation tasks. By crafting ``textbook quality" data we were able to train a model that surpasses almost all open-source models on coding benchmarks such as HumanEval and MBPP despite being 10x smaller in model size and 100x smaller in dataset size. We hypothesize that such high quality data dramatically improves the learning efficiency of language models for code as they provide clear, self-contained, instructive, and balanced examples of coding concepts and skills.

There remains a number of limitations of our model compared to larger models for code.  Firstly, \phione is specialized in Python coding, which restricts its versatility compared to multi-language models. Secondly, \phione lacks the domain-specific knowledge of larger models such as programming with specific APIs or using less common packages. Lastly, due to the structured nature of the datasets and the lack of diversity in terms of language and style, \phione is less robust to stylistic variations or errors in the prompt (for instance, its performance substantially degrades when there are grammatical mistakes in the prompt). We expand on these limitations and give examples of the failure modes of \phione in Appendix~\ref{app:limitations}. 

None of these limitations seem fundamental, and with more work our approach could be used to tackle each one of them, although it is unclear what scaling might be necessary to overcome them (both for the model size and the dataset size). We also believe that significant gains could be achieved by using GPT-4 to generate the synthetic data instead of GPT-3.5, as we noticed that GPT-3.5 data has a high error rate. It is interesting that \phione is able to achieve such high coding proficiency despite those errors (a similar phenomenon was observed in \cite{allen2023physics} where a language model can be trained on data with 100\% error rate and still generate correct answers at test time).

More generally, our work provides evidence that developing good methodology for creating high-quality datasets is a central direction of research for advancing natural language processing and related fields (see also \cite{jung2023impossible} for further evidence). However, creating high-quality datasets is not a trivial task, and it poses several challenges that need to be addressed. One challenge is to ensure that the dataset covers all the relevant content and concepts that one wants the model to learn, and that it does so in a balanced and representative way. Another challenge is to ensure that the dataset is truly diverse and non-repetitive, so that the model does not simply overfit to the data or memorize specific patterns or solutions. This requires finding ways to inject randomness and creativity into the data generation process, while still maintaining the quality and the coherence of the examples. Moreover, even after creating such datasets, we lack a good methodology to measure and evaluate the amount of diversity and redundancy in the data. For example, if we have a dataset with coding exercises, it is hard to determine how many different variations of each exercise exist, and how they are distributed across the dataset. Finally, as language models themselves will be used to curate data for future language models, it further increases the urgency on the ethical and social implications of training such models, such as the accountability, the transparency, and the bias of the data and the models that are involved in this process.

\newpage
\bibliographystyle{alpha}
\bibliography{mainbib}

\appendix \clearpage
\section{Additional examples for Section \ref{sec:sparks}} \label{sec:additional_example}

In Section~\ref{sec:sparks}, we discussed with a few examples on how finetuned modes shows a substantial improvement over the base model in executing tasks that are \textit{not} featured in the finetuning dataset. Here we provide additional examples and details to further illustrate improvements. %
\subsection{Finetuning improves the model's understanding ability}

The example below is challenging due to the prevalence of various logical operators such as ``and" and ``not" that link numerous statements. Prior to finetuning, the model is confused by the logic embedded in the statement. It struggles to differentiate between ``elements" and ``indices", and is unsure about which array to use. Conversely, the finetuned model is adept at accurately identifying all relevant arrays and indices, while correctly implementing the logical operators.

\begin{AIbox}{}
\textbf{Prompt:}\\\tt{\footnotesize """ Input is an array A, an array B, an array C, an array E, a numpy matrix D. I want to find all the indices of D whose elements are in the union of B and C and whose row index is not in A and column index is in D. Loop through the elements in D. Return the list of indices.\newline"""}
\\
\begin{minipage}[t]{0.37\linewidth}
\textbf{\phione}
\begin{lstlisting}[language=Python, style=mystyle]
indices = []
for i in range(D.shape[0]):
    for j in range(D.shape[1]):
        if D[i][j] in B or D[i][j] in C:
            if i not in A and j in D:
                indices.append((i,j))
return indices
\end{lstlisting}
\end{minipage}
\begin{minipage}[t]{0.29\linewidth}
\textbf{\phionebase}
\begin{lstlisting}[language=Python, style=mystyle]
indices = []
for i in range(len(D)):
    for j in range(len(D[i])):
        if D[i][j] in B and D[i][j] not in A:
            indices.append(j)
return indices
\end{lstlisting}
\end{minipage}
\begin{minipage}[t]{0.31\linewidth}
\textbf{\phionesmall}
\begin{lstlisting}[language=Python, style=mystyle]
indices = []
for i in range(D.shape[1]):
    if D[0][i] in B and D[0][i] not in A and D[i][0] in C and D[i][0] not in E:
        indices.append(i)
return indices
\end{lstlisting}
\end{minipage}
\end{AIbox}

\subsection{Finetuning improves the model's ability to use external libraries}

\paragraph{Pytorch example.} In this PyTorch example, we ask the model to write a  modified gradient update:

\begin{AIbox}{}
\textbf{Prompt:}\\ \tt{\footnotesize """ Write a pytorch function. The input is a model, take the gradient of each parameter of the model, do an SVD of each gradient to get U, Sigma, V, take the top k singular values of Sigma as Sigmak, then do a update step to the parameter using Uk*Sigmak*Vk times the norm of the parameter.\newline"""}
\\
\begin{minipage}[t]{0.38\linewidth}
\textbf{\phione}
\begin{lstlisting}[language=Python, style=mystyle]
...
for p in model.parameters():
    if p.grad is not None:
        U, Sigma, V = p.grad.svd()
        Sigmak = Sigma[:k].diag()
        p.data.add_(torch.mm(U[:,:k], torch.mm(Sigmak, V[:k,:])).mul_(norm))
\end{lstlisting}
\end{minipage}
\begin{minipage}[t]{0.24\linewidth}
\textbf{\phionebase}
\begin{lstlisting}[language=Python, style=mystyle]
...
self.model = model
self.k = k
self.eps = eps
self.max_iter = max_iter
self.verbose = verbose
self.svd_u = None
self.svd_sigma = None
self.svd_v = None
self.svd_k = None
self.svd_u_k = None
...
\end{lstlisting}
\end{minipage}
\begin{minipage}[t]{0.36\linewidth}
\textbf{\phionesmall}
\begin{lstlisting}[language=Python, style=mystyle]
def svd(x):
    u, s, v = np.linalg.svd(x)
    return u, s, v
def update_parameter(parameter, grad):
    u, s, v = svd(grad)
    parameter = u[:, :k] @ np.diag(s[:k]) @ v[:k, :]
    parameter += np.diag(sigmoid(parameter))
    return parameter
\end{lstlisting}
\end{minipage}
\end{AIbox}

The code by \phionesmall reveals a lack api usage and limitations in understanding, leading to multiple significant missteps. First, it uses \texttt{np.linalg.svd(x)}, which is incompatible with PyTorch tensors. Second, it erroneously assigns the SVD result directly to the parameter, deviating from the prompt's instruction to add it. On the other hand, \phionebase is completely clueless and only produces a sequence of meaningless definitions. This indicates that improving the model's logical reasoning ability (even only on simple functions) can greatly enhance the model's general coding ability.

\paragraph{Pyplot example.} The final API example is a Pyplot application. The model is generally good with Pyplot, so we design a challenging task to ask the model implement an \textbf{animation}.

\begin{AIbox}{}
\textbf{Prompt:}\\
\tt{\footnotesize """Create an x array that are integers from $1$ to $20$, $y$ array that is equal to the elementwise 
$\cos$ of x divided by element-wise sqrt(x + 1). Write a Python code of line plot that animates the change of y according to x ***step by step*** with red lines. However, during the animation, when x[i] is larger than 11, change the animation line color from red to green.\newline"""}
\\
\begin{minipage}[t]{0.33\linewidth}
\textbf{\phione}
\begin{lstlisting}[language=Python, style=mystyle]
...
x = np.arange(1, 21)
y = np.cos(x) / np.sqrt(x + 1)
fig, ax = plt.subplots()
line, = ax.plot([], [], color='red')
...
ax.set_title('Animation of Cosine and Cosine-Square-Root')
def animate(i):
    if x[i] > 11:
        line.set_color('green')
    else:
        line.set_color('red')
    line.set_data(x[:i+1], y[:i+1])
    return line,
anim = animation.FuncAnimation(fig, animate, frames=len(x), interval=500, blit=True)
plt.show()
\end{lstlisting}
\end{minipage}
\begin{minipage}[t]{0.33\linewidth}
\textbf{\phionebase}
\begin{lstlisting}[language=Python, style=mystyle]
...
fig, ax = plt.subplots()
x = np.arange(1, 21)
y = np.cos(x) / np.sqrt(x + 1)
line, = ax.plot([], [], lw=2)
def init():
    line.set_data([], [])
    return line,
def animate(i):
    if x[i] > 11:
        line.set_color('green')
    else:
        line.set_color('red')
    return line,
anim = FuncAnimation(fig, animate, init_func=init, frames=len(x), interval=50, blit=True)
\end{lstlisting}
\end{minipage}
\begin{minipage}[t]{0.33\linewidth}
\textbf{\phionesmall}
\begin{lstlisting}[language=Python, style=mystyle]
...
fig, ax = plt.subplots()
x_array = list(range(1, 21))
y_array = [math.cos(xi/10) for xi in x_array]
def animate(i):
    ...
    ax.plot(x_array, y_array)
    if x_array[i] > 11:
        color = "red"
    else:
        color = "green"
    ax.scatter(x_array[i], y_array[i], color=color)
    return ax
ani = animation.FuncAnimation(fig, animate, frames=len(x_array), interval=50, blit=True)
plt.show()
\end{lstlisting}
\end{minipage}
\end{AIbox}

\phionesmall produces a very incorrect completion. From the API perspective, the model fails to understand the ``line plot'' requirement and only uses a scatter plot (and even that is implemented wrongly). \phionebase gets the template correct, but misses the core function to update the line plot every iteration. 

\section{Limitation of \phione} %
\label{app:limitations}
While finetuning through simple exercises significantly enhances the model's overall performance, there are certain constraints intrinsic to our model that cannot be overcome solely by finetuning. Firstly, our model has only 1.3B parameters trained with only 7B tokens, this restricts our model's capacity to manage more complex tasks such as developing an intricate Flask application, in comparison to other models like Starcoder. Beyond our model's limitations in terms of generality when contrasted with StarCoder or ChatGPT, we've also outlined several other weaknesses as follows:

\paragraph{Sensitivity to prompt variations.} Our model is sensitive to various perturbations of prompts. First, its performance drops significantly as the length of the prompt increases, as it tends to ignore, forget or misinterpret parts of the prompt when it is too long. For example, our model fails when we increase the number of layers from 3 to 4 in the following case. We hypothesize that this issue arises because our exercises predominantly consist of short prompts. Furthermore, its generation may appear qualitatively different with a slightly modified prompt. In this case, with an additional \texttt{import torch} command, the model tends to succeed on the very task that it failed previously.

\begin{AIbox}{}
\begin{tabular}{p{0.29\textwidth}p{0.29\textwidth}p{0.36\textwidth}} 
\textbf{\small Prompt:}\newline
\tt{\scriptsize  """
Create a neural network class in Pytorch with 3 layers. Implement a new "rr" method to this new function: It takes input $x$. Apply the first layer on $x$ and assign the output to $y$, then apply the second layer on $y$ and assign the output to $z$, then apply the third layer on $z$ and assign it to $w$, then sum up $y, z, w$ and assign it to $Q$. Return $Q$.
\newline
"""
}
& 
\textbf{\small Prompt:}\newline
\tt{\scriptsize """ 
Create a neural network class in Pytorch with {\color{red}4} layers. Implement a new "rr" method to this new function: It takes input $x$. Apply the first layer on $x$ and assign the output to $y$, then apply the second layer on $y$ and assign the output to $z$, then apply the third layer on $z$ and assign it to $w$, {\color{red} then apply the forth layer on $w$ and assign it to $p$, then sum up $y, z, w, p$ and assign it to $Q$}. Return $Q$. 
\newline 
"""
}
&
\textbf{\small Prompt:}
\begin{lstlisting}[language=Python, style=mystyle]
import torch
\end{lstlisting}
\vspace{-0.2cm}
\tt{\scriptsize """ Create a neural network class in Pytorch with {\color{red}4} layers. Implement a new "rr" method to this new function: It takes input $x$. Apply the first layer on $x$ and assign the output to $y$, then apply the second layer on $y$ and assign the output to $z$, then apply the third layer on $z$ and assign it to $w$, {\color{red} then apply the forth layer on $w$ and assign it to $p$, then sum up $y, z, w, p$ and assign it to $Q$}. Return $Q$.
\newline
\texttt{"""}}
\\ \hdashline
\begin{lstlisting}[language=Python, style=mystyle]
class NeuralNetwork(nn.Module):
    def __init__(self):
        super(NeuralNetwork, self).__init__()
        self.layer1 = nn.Linear(2, 5)
        self.layer2 = nn.Linear(5, 3)
        self.layer3 = nn.Linear(3, 1)
        
    def rr(self, x):
        y = self.layer1(x)
        z = self.layer2(y)
        w = self.layer3(z)
        Q = y + z + w
        return Q
\end{lstlisting}
&
\begin{lstlisting}[language=Python, style=mystyle]
class NeuralNetwork:
    def __init__(self):
        self.layers = []
    
    def add_layer(self, layer):
        self.layers.append(layer) 
    ...
    def rr(self, x):
        y = self.forward(x)
        z = self.forward(y)
        w = self.forward(z)
        p = self.forward(w)
        Q = y + z + w + p
        return Q
\end{lstlisting}
&
\begin{lstlisting}[language=Python, style=mystyle]
class NeuralNetwork(torch.nn.Module):
    def __init__(self):
        super(NeuralNetwork, self).__init__()
        self.layer1 = torch.nn.Linear(10, 20)
        ...

    def forward(self, x):
        x = torch.nn.functional.relu(self.layer1(x))
        y = torch.nn.functional.relu(self.layer2(x))
        z = torch.nn.functional.relu(self.layer3(y))
        w = torch.nn.functional.relu(self.layer4(z))
        p = torch.nn.functional.relu(self.output(w))
        Q = p + y + z + w
        return Q
\end{lstlisting}
\end{tabular}\vspace{-1.5em}
\end{AIbox}
\clearpage
\paragraph{Sensitivity to natural language inputs.} \phione demonstrates less robustness in handling natural language compared to ChatGPT or StarCoder, particularly with ambiguous prompts. This may be because we filter out certain types of data from the training process to guarantee textbook-level quality. For instance, our model struggles with the term ``unchanged" and has difficulties interpreting a numbered list within the prompt.
\begin{AIbox}{}
\begin{tabular}{p{0.33\textwidth}p{0.33\textwidth}p{0.28\textwidth}} 
\textbf{Prompt:}\newline\tt{\scriptsize """ Alice is moving a point along x,y axes at every iteration. Every time Alice moves the point randomly with one of the options: (1). x increases by 2, y decreases by 5; (2). x increases by 1, y decreases by 2, (3). \textcolor{red}{x stays unchanged}, y increase by 4. Simulate the movement of the points.\newline"""}
& 
\textbf{Prompt:}\newline \tt{\scriptsize """ Alice is moving a point along x,y axes at every iteration. Every time Alice moves the point randomly with one of the options: (1). x increases by 2, y decreases by 5; (2). x increases by 1, y decreases by 2, (3). \textcolor{red}{x stays at x}, y increase by 4. Simulate the movement of the points.\newline"""} 
& 
\textbf{Prompt:} \newline \tt{\scriptsize """ Alice is moving a point along x,y axes at every iteration. Every time Alice moves the point randomly with one of the options: (1). move by (2, -5); (2). move by (1, 2); (3). move by (0, 4); Simulate the movement of the points.\newline"""}\\\vspace{-1em}
\begin{lstlisting}[language=Python, style=mystyle]
x, y = 0, 0
for i in range(10):
    move = random.choice([1, 2, 3])
    if move == 1:
        x += 2
        y -= 5
    elif move == 2:
        x += 1
        y -= 2
    elif move == 3:
        x = 0
        y += 4
return f"Alice's final position is ({x}, {y})."
\end{lstlisting}
& \vspace{-1em}
\begin{lstlisting}[language=Python, style=mystyle]
x, y = 0, 0
for i in range(10):
    move = random.choice([1, 2, 3])
    if move == 1:
        x += 2
        y -= 5
    elif move == 2:
        x += 1
        y -= 2
    elif move == 3:
        x = x
        y += 4
    print(f"Alice moved {move} point to ({x}, {y})")
\end{lstlisting}
& \vspace{-1em}
\begin{lstlisting}[language=Python, style=mystyle]
x, y = 0, 0
for i in range(n):
    move = random.choice([(1, 2), (2, -5), (1, 2), (3, 0), (0, 4)])
    x += move[0]
    y += move[1]
return f"Alice moved {x} units to the right and {y} units down after {n} iterations."
\end{lstlisting}
\emph{\scriptsize Note: the model hallucinates the moves (1, 2), (3, 0).}
\end{tabular}\vspace{-1.5em}

\end{AIbox}

\clearpage
\paragraph{Bad at counting and spatial reasoning.} 
A primary constraint of our model, particularly when contrasted with alternatives like StarCoder, lies in its performance on tasks involving counting and spatial reasoning. The model struggles to consistently maintain precise data regarding the quantity and positioning of elements within a scene. To illustrate, consider the following example:

\begin{AIbox}{}
\begin{tabular}[H]{ m{0.48\textwidth}m{0.48\textwidth}} \\
\multicolumn{2}{p{0.96\textwidth}}{\textbf{Prompt:}\newline\tt{
\footnotesize """ Generate a tkinter application, I want two textfields side by side at the top, three buttons at the bottom, one named $A$ on the left, one named $B$ on the right, one named $C$ at the middle.
\newline"""}
} \\
\ \vspace{-0.5em} \\
\phione & \phionesmall\\
\begin{lstlisting}[language=Python, style=mystyle]
...
text_a = tk.Text(root, width=30, height=5)
text_b = tk.Text(root, width=30, height=5)
text_c = tk.Text(root, width=30, height=5)

# Create the buttons
button_a = tk.Button(root, text="A", width=10, height=2)
button_b = tk.Button(root, text="B", width=10, height=2)
button_c = tk.Button(root, text="C", width=10, height=2)

# Place the text fields and buttons in the top row
text_a.grid(row=0, column=0, padx=5, pady=5)
text_b.grid(row=0, column=1, padx=5, pady=5)
text_c.grid(row=0, column=2, padx=5, pady=5)

# Place the buttons in the bottom row
button_a.grid(row=1, column=0, columnspan=3, padx=5, pady=5)
button_b.grid(row=1, column=1, columnspan=3, padx=5, pady=5)
button_c.grid(row=1, column=2, columnspan=3, padx=5, pady=5)
...
\end{lstlisting}
&
\begin{lstlisting}[language=Python, style=mystyle]
window = tk.Tk()
window.title("TextFields")
# Create the left and right textfields
text1 = tk.Text(window, text="A")
text1.pack()
text2 = tk.Text(window, text="B")
text2.pack()
# Create the middle button
button = tk.Button(window, text="C")
button.pack()
# Start the tkinter application
window.mainloop()
\end{lstlisting}
\phionebase
\begin{lstlisting}[language=PythonPlus, style=mystyle]
root = tk.Tk()
root.title("Text Fields")
root.geometry("300x200")
root.resizable(False, False)
root.configure(background="white")
root.configure(font=("Arial", 12))
root.configure(foreground="black")
root.configure(relief="groove")
root.configure(padx=10, pady=10)
root.configure(state="disabled")
# The model keeps outputting nonsensical code.
\end{lstlisting}
\end{tabular}\vspace{-1.5em}
\end{AIbox}

Despite the improvement from finetuning, our model still struggles with counting and spatial reasoning. It generates an extra textfield and misplaces the button in the scene.

\clearpage
\section{Examples for Section \ref{sec:contaim}}\label{sec:additional_example_contam}
In this section, we provide example pairs of codes captured with different AST match rates. Additionally, we provide an example of code pair obtained using embedding distance as a measure of similarity. %

\vspace{-0.5em}
\paragraph{AST match rate = 1.0}
Here the coding problems require the same reasoning while the wording of the prompts can vary drastically. Particularly, the prompt uses a real-world event, i.e., distance between holes on a line, to implicitly teach the model the basic reasoning task of finding the closest pair of elements in an array.
\begin{AIbox}{}\vspace{-1em}
\begin{tabular}[H]{m{0.49\textwidth}m{0.49\textwidth}} \\
\multicolumn{1}{c}{\textbf{HumanEval}}  & \multicolumn{1}{c}{\textbf{CodeExercises}}\\ 
\begin{lstlisting}[language=Python, style=mystyle]
from typing import List, Tuple
def find_closest_elements(numbers: List[float]) -> Tuple[float, float]:
    """ From a supplied list of numbers (of length at least two) select and return two that are the closest to each
    other and return them in order (smaller number, larger number).
    >>> find_closest_elements([1.0, 2.0, 3.0, 4.0, 5.0, 2.2])
    (2.0, 2.2)
    >>> find_closest_elements([1.0, 2.0, 3.0, 4.0, 5.0, 2.0])
    (2.0, 2.0)
    """
    numbers.sort()
    min_diff = float('inf')
    closest_pair = None
    for i in range(len(numbers) - 1):
        diff = numbers[i+1] - numbers[i]
        if diff < min_diff:
            min_diff = diff
            closest_pair = (numbers[i], numbers[i+1])
    return closest_pair
\end{lstlisting} &
\begin{lstlisting}[language=Python, style=mystyle]
from typing import List, Tuple
def find_closest_two_holes(holes: List[int]) -> (int, int):
    """
    Finds the two closest holes on a line.
    Args:
        holes: A list of integers representing the positions of holes on a line.
    Returns:
        A tuple of two integers representing the positions of the two closest holes on the line.
    """
    holes.sort()
    min_distance = float('inf')
    closest_holes = None
    for i in range(len(holes) - 1):
        distance = holes[i+1] - holes[i]
        if distance < min_distance:
            min_distance = distance
            closest_holes = (holes[i], holes[i+1])
    return closest_holes
\end{lstlisting}
\end{tabular}\vspace{-1.5em}
\end{AIbox}
\clearpage

\paragraph{AST match rate = 0.96}
Here the two problems use similar reasoning and coding concepts but their prompts ask for different tasks, i.e., returning a pair of numbers versus computing their average.

\vspace{-0.5em}
\begin{AIbox}{}\vspace{-1em}
\begin{tabular}[H]{m{0.48\textwidth}m{0.49\textwidth}} \\
\multicolumn{1}{c}{\textbf{HumanEval}}  & \multicolumn{1}{c}{\textbf{CodeExercises}}\\
\begin{lstlisting}[language=Python, style=mystyle]
from typing import List
def find_closest_elements(numbers: List[float]) -> Tuple[float, float]:
    """ From a supplied list of numbers (of length at least two) select and return two that are the closest to each
    other and return them in order (smaller number, larger number).
    >>> find_closest_elements([1.0, 2.0, 3.0, 4.0, 5.0, 2.2])
    (2.0, 2.2)
    >>> find_closest_elements([1.0, 2.0, 3.0, 4.0, 5.0, 2.0])
    (2.0, 2.0)
    """
    numbers.sort()
    min_diff = float('inf')
    closest_pair = None
    for i in range(len(numbers) - 1):
        diff = numbers[i+1] - numbers[i]
        if diff < min_diff:
            min_diff = diff
            closest_pair = (numbers[i], numbers[i+1])
    return closest_pair
\end{lstlisting} &
\begin{lstlisting}[language=Python, style=mystyle]
from typing import List
def find_closest_pair_average(li: List[int]) -> float:
    """
    Returns the average of the two integers in the list that are closest to each other.
    If there are multiple pairs with the same minimum difference, the function returns the average of the first pair it encounters.
    Args:
    - li: a list of integers
    Returns:
    - a float representing the average of the two integers in the list that are closest to each other
    """
    li.sort()
    min_diff = float('inf')
    closest_pair = None
    for i in range(len(li)-1):
        diff = li[i+1] - li[i]
        if diff < min_diff:
            min_diff = diff
            closest_pair = (li[i], li[i+1])
    return sum(closest_pair) / 2
\end{lstlisting}
\end{tabular}\vspace{-1.5em}
\end{AIbox}

\paragraph{AST match rate $\mathbf{\leq}$ 0.9}
When the AST match rate $\leq 0.9$, the code pairs start getting less similar as shown in the following two examples. Here, the AST match rate is $0.9$ and $0.83$, respectively.

\begin{AIbox}{}\vspace{-1em}
\begin{tabular}[H]{m{0.48\textwidth}m{0.49\textwidth}} \\
\multicolumn{1}{c}{\textbf{HumanEval}}  & \multicolumn{1}{c}{\textbf{CodeExercises}}\\ 
\begin{lstlisting}[language=Python, style=mystyle]
from typing import List
def all_prefixes(string: str) -> List[str]:
    """ Return list of all prefixes from shortest to longest of the input string
    >>> all_prefixes('abc')
    ['a', 'ab', 'abc']
    """

    prefixes = []
    for i in range(len(string)):
        prefixes.append(string[:i+1])
    return prefixes
\end{lstlisting} &
\begin{lstlisting}[language=Python, style=mystyle]
from typing import List
def increase_after_each(li: List[int]) -> List[int]:
    """
    Given a list of integers, returns a new list where each element is increased by the number of elements in the original list that come after it.
    Args:
    - li: A list of integers
    Returns:
    - A new list of integers where each element is increased by the number of elements in the original list that come after it.
    """
    result = []
    for i in range(len(li)):
        result.append(li[i] + len(li[i+1:]))
    return result
\end{lstlisting}
\end{tabular}\vspace{-1.5em}
\end{AIbox}

\begin{AIbox}{}\vspace{-.5em}
\begin{tabular}[H]{ m{0.48\textwidth}m{0.49\textwidth}} \\
\multicolumn{1}{c}{\textbf{HumanEval}}  & \multicolumn{1}{c}{\textbf{CodeExercises}}\\
\begin{lstlisting}[language=Python, style=mystyle]
from typing import List
def rescale_to_unit(numbers: List[float]) -> List[float]:
    """ Given list of numbers (of at least two 
    elements), apply a linear transform to that 
    list, such that the smallest number will 
    become 0 and the largest will become 1
    >>> rescale_to_unit([1.0, 2.0, 3.0, 4.0, 5.0])
    [0.0, 0.25, 0.5, 0.75, 1.0]
    """
    min_num = min(numbers)
    max_num = max(numbers)
    return [(num - min_num) / (max_num - min_num) for num in numbers]
\end{lstlisting} &
\begin{lstlisting}[language=Python, style=mystyle]
from typing import List
def frequency_ranges_plot(values: List[float]) -> List[List[int]]:
    """
    Returns a list of lists where each inner list represents a frequency range and contains the count of values
    that fall within that range. The ranges are determined by dividing the range of values into 10 equal parts.
    Args:
    - values: A list of float values
    Returns:
    - A list of lists where each inner list contains two integers representing the lower and upper bounds of the range
    and the count of values that fall within that range.
    """
    min_val = min(values)
    max_val = max(values)
    range_size = (max_val - min_val) / 10
    ranges = [[min_val + i*range_size, min_val + (i+1)*range_size, 0] for i in range(10)]
    
    for val in values:
        for i, r in enumerate(ranges):
            if r[0] <= val < r[1]:
                ranges[i][2] += 1
                break  
    return [[int(r[0]), int(r[1]), r[2]] for r in ranges]
\end{lstlisting}
\end{tabular}\vspace{-1.5em}
\end{AIbox}

\paragraph{Embedding Distance = 0.16}
Here the two problems have similar Python Docstrings, function names, as
well as the code structure which can be extracted with using the L2 distance between the normalized CodeGen-Mono 350M embedding for each of them.
\begin{AIbox}{}\vspace{-1em}
\begin{tabular}[H]{m{0.49\textwidth}m{0.49\textwidth}} \\
\multicolumn{1}{c}{\textbf{HumanEval}}  & \multicolumn{1}{c}{\textbf{CodeExercises}}\\ 
\begin{lstlisting}[language=Python, style=mystyle]
def sum_product(numbers: List[int]) -> Tuple[int, int]:
    """ For a given list of integers, return a 
    tuple consisting of a sum and a product of 
    all the integers in a list.
    Empty sum should be equal to 0 and empty 
    product should be equal to 1.
    >>> sum_product([])
    (0, 1)
    >>> sum_product([1, 2, 3, 4])
    (10, 24)
    """

    sum_value = 0
    prod_value = 1

    for n in numbers:
        sum_value += n
        prod_value *= n
    return sum_value, prod_value
\end{lstlisting} &
\begin{lstlisting}[language=Python, style=mystyle]
from typing import List, Tuple
def all_numbers_sum_product(numbers: List[int]) -> Tuple[int,int]:
    """
    Returns a tuple containing the sum and 
    product of all the numbers in the input list.
    Args:
    - numbers (List[int]): a list of integers
    Returns:
    - a tuple containing two integers:
        - the sum of all the numbers in the input list
        - the product of all the numbers in the input list
    """
    sum_of_numbers = 0
    product_of_numbers = 1
    for num in numbers:
        sum_of_numbers += num
        product_of_numbers *= num
    return (sum_of_numbers, product_of_numbers)
\end{lstlisting}
\end{tabular}\vspace{-1.5em}
\end{AIbox}

\end{document}